\documentclass[10pt,twocolumn,letterpaper]{article}

\usepackage[pagenumbers]{cvpr} 

\usepackage{times}
\usepackage{epsfig}
\usepackage{graphicx}
\usepackage{amsmath}
\usepackage{amssymb}
\usepackage[table]{xcolor}
\usepackage{booktabs}
\usepackage{tabularx}
\usepackage{gensymb}
\usepackage{enumitem}
\usepackage{multirow}
\usepackage{makecell}
\usepackage{caption}
\usepackage{subcaption}
\definecolor{cvprblue}{rgb}{0.21,0.49,0.74}

\usepackage{xurl}

\usepackage[pagebackref,breaklinks,colorlinks,citecolor=cvprblue]{hyperref}

\def\paperID{10974} 
\def\confName{CVPR}
\def\confYear{2024}

\title{Multi-modal learning for geospatial vegetation forecasting} 

\author{Vitus Benson\textsuperscript{1,2,3,*} \and Claire Robin\textsuperscript{1,2} \and Christian Requena-Mesa\textsuperscript{1,2} \and Lazaro Alonso\textsuperscript{1} \and Nuno Carvalhais\textsuperscript{1,2} \and José Cortés\textsuperscript{1} \and Zhihan Gao\textsuperscript{4} \and Nora Linscheid\textsuperscript{1} \and Mélanie Weynants\textsuperscript{1} \and Markus Reichstein\textsuperscript{1,2} \\
\textsuperscript{1} Max-Planck-Institute for Biogeochemistry  \quad \textsuperscript{2} ELLIS Unit Jena  \quad \textsuperscript{3} ETH Zürich  \\ \textsuperscript{4} Hong Kong University of Science and Technology  \quad \textsuperscript{*} {\tt vbenson@bgc-jena.mpg.de}
}

\begin{document}
\maketitle

\begin{abstract}
    The innovative application of precise geospatial vegetation forecasting holds immense potential across diverse sectors, including agriculture, forestry, humanitarian aid, and carbon accounting. To leverage the vast availability of satellite imagery for this task, various works have applied deep neural networks for predicting multispectral images in photorealistic quality. However, the important area of vegetation dynamics has not been thoroughly explored. Our study breaks new ground by introducing GreenEarthNet, the first dataset specifically designed for high-resolution vegetation forecasting, and Contextformer, a novel deep learning approach for predicting vegetation greenness from Sentinel 2 satellite images with fine resolution across Europe. Our multi-modal transformer model Contextformer leverages spatial context through a vision backbone and predicts the temporal dynamics on local context patches incorporating meteorological time series in a parameter-efficient manner. The GreenEarthNet dataset features a learned cloud mask and an appropriate evaluation scheme for vegetation modeling. It also maintains compatibility with the existing satellite imagery forecasting dataset EarthNet2021, enabling cross-dataset model comparisons. Our extensive qualitative and quantitative analyses reveal that our methods outperform a broad range of baseline techniques. This includes surpassing previous state-of-the-art models on EarthNet2021, as well as adapted models from time series forecasting and video prediction. To the best of our knowledge, this work presents the first models for continental-scale vegetation modeling at fine resolution able to capture anomalies beyond the seasonal cycle, thereby paving the way for predicting vegetation health and behaviour in response to climate variability and extremes. We provide open source code and pre-trained weights to reproduce our experimental results under \url{https://github.com/vitusbenson/greenearthnet} \cite{benson_2024_10793870}.

\end{abstract}

\begin{figure}[t]
    \centering
    \includegraphics[width=\columnwidth]{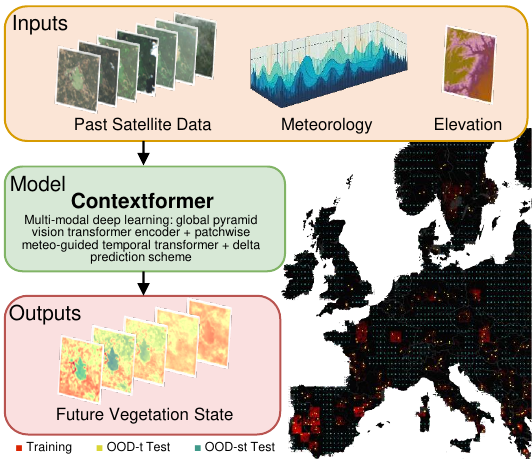}  
    \caption{Future vegetation status $\hat{V}$ is predicted with deep learning models $f$ from past satellite imagery $X$, past and future weather $C$ and elevation $E$. The dataset GreenEarthNet spans across Europe with minicubes split into train (red markers), temporal OOD test (ood-t, yellow) and spatio-temporal OOD test (ood-st, blue) subsets.}
    \label{fig:task}
\end{figure}

\section{Introduction}

Optical satellite images have been proven useful for monitoring vegetation status. This is essential for a variety of applications in agricultural planning, forestry advisory, humanitarian assistance or carbon monitoring. In all these cases, prognostic information is relevant: Farmers want to know how their farmland may react to a given weather scenario \cite{wolanin.etal_2019}. Humanitarian organisations need to understand the localized impact of droughts on pastoral communities for mitigation of famine with anticipatory action \cite{meshesha.etal_2020}. Afforestation efforts need to consider how their forests react to future climate \cite{sturm.etal_2022}. However, providing such prognostic information through fine resolution vegetation forecasts is challenging as it requires a model that considers ecological memory effects \cite{kraft.etal_2019}, spatial interactions and the influence of weather variations. Deep neural networks have proven successful at modeling relationships in space, time or across modalities. Hence, their application to vegetation forecasting given a sufficiently large dataset seems natural.

So far, deep learning in the domain of vegetation forecasting can be roughly grouped into two categories: low-resolution global vegetation forecasting and high-resolution local satellite imagery forecasting. The former \cite{ji.peters_2004,kraft.etal_2019, barrett.etal_2020, lees.etal_2022, zeng.etal_2022, shamseddin.gall_2023, martinuzzi.etal_2023} builds upon long-term measurements of vegetation status from the coarse resolution AVHRR and MODIS satellites. These methods overlook the heterogeneity within each pixel, e.g. a grassland will react very different than a forest, two neighbouring fields can have almost opposite dynamics depending on the type of crops (\textit{e.g.} winter wheat vs. maize), and vegetation on a north-facing slope close to a river will generally react different to drought stress than on a rocky south-facing slope. The latter \cite{requena-mesa.etal_2021, kladny2024enhanced, diaconu.etal_2022, robin.etal_2022, smith.etal_2023, tseng.etal_2023}, in contrast, aims at modeling the field-scale heterogeneity as observed from the Sentinel 2 and Landsat satellites through self-supervised learning. However, these approaches have so far focused on perceptual image quality instead of vegetation dynamics, which renders their suitability for vegetation forecasting uncertain. 


The largest dataset for high resolution vegetation forecasting \cite{xiong.etal_2022} is called EarthNet2021 \cite{requena-mesa.etal_2021}. In theory, a model trained on EarthNet2021 can forecast satellite images of high perceptual quality. It is, however, harder to assess if this skill also translates to predicting derived vegetation dynamics. Here, the suitability of EarthNet2021 is limited by a faulty cloud mask, insufficient baselines and a poorly interpretable evaluation protocol. For instance, natural vegetation follows a strong seasonal cycle, making a vegetation climatology a necessary baseline to compare with, which is lacking on EarthNet2021. 

In this paper, we approach continental-scale modeling of vegetation dynamics. To achieve this, we predict remotely sensed vegetation greenness at 20m conditioned on coarse-scale weather. For this, we introduce the \emph{GreenEarthNet} dataset. It includes Sentinel 2 bands red, green, blue and near-infrared and a high quality deep learning-based cloud mask, which allows to distinguish between anomalies due to data corruption and those due to meteorological and anthropogenic influence. The training locations and spatial and temporal dimensions of GreenEarthNet are kept consistent with the EarthNet2021 dataset, which enables the re-use of the leading models ConvLSTM \cite{diaconu.etal_2022}, SGED-ConvLSTM \cite{kladny2024enhanced} and Earthformer \cite{gao.etal_2022} as baselines for vegetation forecasting. In other words, GreenEarthNet is a complete remake of the EarthNet2021 dataset, removing all its weaknesses and enabling self-supervised learning for geospatial vegetation forecasting. To advance state-of-the-art on the new dataset, we present a light-weight transformer model: the \emph{Contextformer}. It utilizes a Pyramid Vision Transformer \cite{wang.etal_2021a, wang.etal_2022c} as vision backbone and local context patches to make use of spatial interactions. It then models temporal dynamics conditioned on coarse-scale meteorology with a temporal transformer encoder. Finally, it puts a strong prior on persistence through a delta-prediction scheme starting from an initial, gap filled observation. Fig.~\ref{fig:task} presents a sketch of our GreenEarthNet approach: Future vegetation state ($\hat{V}$) is predicted from past satellite image spectra ($X$), past and future weather data ($C$), and elevation information ($E$) via a deep neural network: the Contextformer ($f$).

Our major \textbf{contributions} can be summarized as follows. \textbf{(1)} We present the GreenEarthNet dataset, the first large-scale dataset suitable for  prediction of within-year vegetation dynamics, including a learned cloud mask and a new evaluation scheme. \textbf{(2)} We introduce the Contextformer, a novel multi-modal transformer model suitable for vegetation forecasting, leveraging spatial context through its vision backbone, and forecasting the temporal evolution of small context patches with a temporal transformer. \textbf{(3)} We compare the Contextformer against a previously unseen variety of state-of-the-art models from related tasks and find it outperforms all of them both across metrics.


\begin{figure*}[t]
    \centering
    \includegraphics[width=\textwidth]{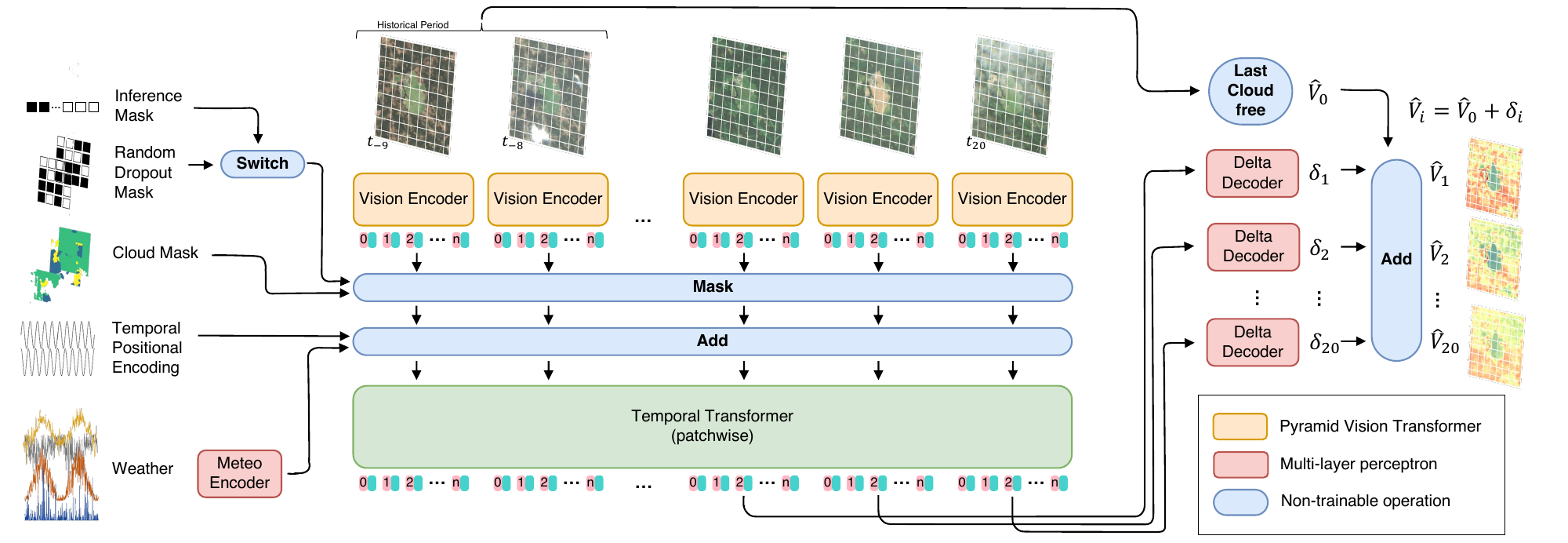}
    \caption{The architecture of our proposed Contextformer.}
    \label{fig:contextformer}
\end{figure*}
\section{Related Work}

\textbf{Vegetation forecasting}
There is a growing interest in vegetation growth forecasting driven by the democratization of machine learning techniques, the availability of remote sensing data, and the urgency to address climate change \cite{ferchichi.etal_2022, kang2016forecasting, cui2020forecasting, lees2022deep}. Numerous studies in vegetation modeling use coarse resolution data from satellites like AVHRR or MODIS \cite{ji.peters_2004, kraft.etal_2019, lees.etal_2022, zeng.etal_2022, martinuzzi.etal_2023}. 
Since 2015, Sentinel-2 has provided high-resolution satellite imagery (up to $10$m), enabling more localized modeling. The introduction of EarthNet2021 \cite{requena-mesa.etal_2021} marked the first dataset for self-supervised Earth surface forecasting, which contains predicting satellite imagery and derived vegetation state with a focus on perceptual quality. Subsequently, the ConvLSTM model \cite{shi.etal_2015} has been widely used for satellite imagery prediction \cite{diaconu.etal_2022, kladny2024enhanced, ahmad.etal_2023, robin.etal_2022, ma.etal_2022a}, hence we are including it as a baseline.

\textbf{Spatio-temporal learning}
Learning spatio-temporal dynamics (as in the case of vegetation forecasting) is a challenge across many disciplines. Often, temporal dynamics dominate, so local time series models can be effective. For instance in traffic, weather or electricity forecasting, time series models such as LSTM \cite{hochreiter.schmidhuber_1997}, Prophet \cite{taylor.letham_2018}, Autoformer \cite{wu.etal_2021a} or NBeats \cite{zeng.etal_2023} yield useful performance. Still, often spatial interactions are important or at least offer additional predictive capacity. For instance in video prediction, ConvNets \cite{babaeizadeh.etal_2021, gao.etal_2022a}, ConvLSTM \cite{shi.etal_2015}, ConvLSTM successors \cite{wang.etal_2023, wu.etal_2021}, PredRNN \cite{wang.etal_2017, wang.etal_2023}, SimVP \cite{tan.etal_2023} and transformers \cite{gupta.etal_2022a, nash.etal_2022a, gao.etal_2022} have been found skillful. Often, the necessity of modeling the spatial component translates to Earth science: spatio-temporal deep learning is being applied for precipitation nowcasting \cite{ravuri.etal_2021a, shi.etal_2017}, weather forecasting \cite{bi.etal_2022, lam.etal_2022, pathak.etal_2022}, climate projection \cite{nguyen.etal_2023}, and wildfire modeling \cite{kondylatos.etal_2022}. Hence, when evaluating our Contextformer model, we need to do so against strong baselines from video prediction \cite{shi.etal_2015, gao.etal_2022, wang.etal_2023, tan.etal_2023}, as a priori one might expect them to outperform also on vegetation forecasting. However, vegetation forecasting does present some unique challenges, it builds upon multi-modal data fusion and requires capturing across-scale relationships (in time and space), which may prove challenging for existing video prediction models and thus interesting to the computer vision community.


\textbf{Multi-modal transformers for data fusion}
Levering remote sensing data often means multi-modal data fusion. Recently, machine learning methods have shown significant advancements in fusing different satellite sensors compared to traditional approaches \cite{steinhausen2018combining, whyte2018new, audebert2018beyond, dalla2015challenges, li2022deep}. This includes recent work on combining Sentinel 2 and SAR data to impute cloudy Sentinel 2 images \cite{wang.etal_2019, meraner.etal_2020, yang.etal_2022}. Gapfilling vegetation time series could also be done with the models presented in this study, as they leverage meteorology to inform the imputation \cite{stucker.etal_2023}. However, as gapfilling is just done in retrospective, one should rather resort to complementary satellite data like SAR.


Transformers \cite{vaswani.etal_2017} offer a compelling approach to handle multi-modal data \cite{jaegle.etal_2022}. Their efficacy in remote sensing has been shown multiple times \cite{ma2022crossmodal, gao.etal_2022, cong.etal_2022, tseng.etal_2023}. In particular, geospatial foundation models \cite{mendieta.etal_2023, reed.etal_2023, cong.etal_2022, yao.etal_2023, smith.etal_2023, tseng.etal_2023} make use of this approach, often through masked token modeling \cite{he.etal_2022} with Vision Transformers \cite{dosovitskiy.etal_2020}. Our Contextformer follows this line of research, yet in contrast to geospatial foundation models, it is more targeted for vegetation forecasting and only utilizes a pre-trained vision transformer as a vision backbone.

\section{Methods}

\subsection{Task}
We predict the future NDVI, a remote sensing proxy of vegetation state ($V^t \in \mathbb{R}^{H\times W}, t \in [T+1, T+K]$) conditioned on past satellite imagery ($X^t \in \mathbb{R}^{H\times W}, t \in [1, T]$), past and future weather ($C^t \in \mathbb{R}, t \in [1, T+K]$) and static elevation maps ($E \in \mathbb{R}^{H\times W}$). Hence, denoting a model $f(.;\theta)$ with parameters $\theta$, we obtain vegetation forecasts as:
\begin{align}
    \hat{V}^{T+1:T+K} = f(X^{1:T}, C^{1:T+K}, E; \theta)
\end{align}
In this paper most models are deep neural networks, trained with stochastic gradient descent to maximize a Gaussian Likelihood. More specifically, the optimal parameters $\theta^{*}$ are obtained by minimizing the mean squared error over valid pixels $V_{*}^t = V^t \odot M_Q^t \odot M_L$, where $M_Q \in \{0,1\}^{H\times W}$ masks pixels that are cloudy, cloud shadow or snow, $M_L \in \{0,1\}^{H\times W}$ masks pixels that are not cropland, forest, grassland or shrubland and $\odot$ denotes elementwise multiplication. Hence the training objective (leaving out dimensions for simplicity) is
\begin{align}
    \theta^{*} = \underset{\theta}{arg min} \frac{\sum (V - \hat{V})^2 \odot M_Q \odot M_L}{\sum M_Q \odot M_L}
\end{align}
In this work $H=W=128\text{px}$, $T=10$ and $K=20$.


\subsection{Our proposed Contextformer model}\label{sec:model}
To tackle the vegetation forecasting task, we develop the Contextformer, a multi-modal transformer model operating on local spatial context patches (hence the name) and trained with self-supervised learning for predicting vegetation state across Europe. Next to historical satellite imagery, it leverages an elevation map and meteorological data.

\textbf{Overview} Our proposed Contextformer follows an \emph{encode-process-decode} \cite{battaglia.etal_2018a} configuration. Encoders and decoders operate in the spatial domain without temporal fusion, while the processor translates latent features temporally in local context patches. We use two encoders (meteo and vision), a temporal transformer processor and a decoder that predicts the delta from the last cloud free NDVI observation (see Fig.~\ref{fig:contextformer}).

\textbf{Encoders} The meteo encoder (for weather) and the delta decoders are parameterized as multi-layer perceptrons (MLPs) (Fig.~\ref{fig:contextformer} red boxes). For the vision encoder (Fig.~\ref{fig:contextformer} yellow boxes), we follow the MMST-ViT model for crop yield prediction \cite{lin.etal_2023} and use a Pyramid Vision Transformer (PVT) v2 B0 \cite{wang.etal_2021a, wang.etal_2022c}, which is particularly suitable for dense prediction tasks. It divides the images of each time step into patches of $4\times4$ px and then creates an embedding for each of the patches with a global receptive field. In other words, information is exchanged spatially at each time step, but not across time steps. We merge multi-scale features from the different stages of an ImageNet pre-trained PVT v2 B0, upscale them to our patches, concatenate and project. The resulting features for each image stack (satellite \& elevation) contain multi-scale and spatial context information.

\textbf{Masked Token Modeling} During training, we drop out patches in a masked token modeling approach \cite{he.etal_2022} and replace them with a learned masking token. We randomly ($p=0.5$) switch between inference mode, where we drop all patches for time steps $10$ to $30$, and random dropout mode, where we mask $70\%$ of the patches for time steps $3$ to $30$. At test time, we only use inference mode: the model never sees future satellite imagery. In addition, we use the cloud mask to drop every patch with at least one cloudy pixel. After applying the masking, a sinusoidal temporal positional encoding \cite{tseng.etal_2023} and the weather embeddings from the meteo encoder are added to each patch.

\textbf{Processor} The temporal transformer (Fig.~\ref{fig:contextformer} green box) processes patches in parallel, i.e. it exchanges information across the $30$ time steps, but spatially only within each $4\times4$ px patch.The idea here is that for ecosystem processes, spatial context is crucial but does not change dynamically. Therefore, separating spatial and temporal processing enhances efficiency. We maintain a small local context ($4\times4$ px) within the temporal encoder due to Sentinel 2's sub-pixel inaccuracies causing slight pixel shifts over time. This approach significantly reduces the model's memory cost during training (by $16\times$), enabling larger batch sizes. Our temporal transformer is implemented following Presto's transformer encoder \cite{tseng.etal_2023}, which is based on the standard vision transformer \cite{dosovitskiy.etal_2020}.

\textbf{Output} Our Contextformer leverages the persistence within the vegetation dynamics by predicting only a deviation from an initial state. More specifically, we compute the last cloud free NDVI observation from the historical period ($10$ time steps) using the cloud mask and use it as initial prediction $\hat{V}^0$. Then, the delta decoder predicts a deviation $\delta^i$ for each of the future period token embeddings (Fig.~\ref{fig:contextformer} right side). The final NDVI prediction is computed as $\hat{V}^i = \hat{V}^0 + \delta^i$.
A similar delta framework was used in training the SGED-ConvLSTM \cite{kladny2024enhanced}. However, that model predicts only one step ahead in an iterative fashion, predicting the deviation to the previous time step. In our multi-step prediction setting, this would result in a cumulative sum on the outputs, which is undesirable for training gradients.

\subsection{GreenEarthNet Dataset}

We present GreenEarthNet, a tailored dataset for high-resolution geospatial vegetation forecasting. It contains spatio-temporal minicubes \cite{loaiza.etal_2023}, that are a collection of $30$ 5-daily satellite images ($10$ historical, $20$ future), $150$ daily meteorological observations and an elevation map. Spatial dimensions are $128\times 128$px ($2.56\times 2.56$km). To enable cross-dataset model comparisons, we re-use the training locations and predictor dimensions from the EarthNet2021 \cite{requena-mesa.etal_2021} dataset for Earth surface forecasting. 

\begin{table}[]
    \centering
    \begin{tabularx}{\columnwidth}{Xcccc}
    \toprule
    & Works /w & & & \\
    Algorithm & GreenEarthNet & Prec & Rec & F1 \\
    \midrule
    Sen2Cor & Yes	& 0.83 & 0.60 & 0.70 \\
    FMask & No & 0.85 & 0.85 & 0.85\\
    KappaMask & No & 0.74 & 0.88 & 0.81\\
    \arrayrulecolor{black!30}\midrule
    UNet RGBNir & Yes & \underline{0.91} & 0.90 & \underline{0.90}\\
    \emph{UNet+Sen2CorSnow} & Yes & 0.83 & \textbf{0.93} & 0.88\\
    \arrayrulecolor{black!30}\midrule
    UNet 13Bands & No & \textbf{0.94} & \underline{0.92} & \textbf{0.93}\\
    \arrayrulecolor{black}\bottomrule
    \end{tabularx}
    \caption{Precision, recall and F1-score of different Sentinel~2 cloud masking algorithms.}
    \label{tab:cloudmask}
\end{table}

\textbf{Satellite and Meteo Layers} GreenEarthNet  includes Sentinel 2 \cite{louis.etal_2016} satellite bands blue, green, red, and near-infrared at $20$m (consistent with EarthNet2021) and E-OBS \cite{cornes.etal_2018} interpolated meteorological station data, which represents high quality meteorology over Europe \cite{bandhauer2022evaluation}. More specifically, the meteorological drivers wind speed, relative humidity, and shortwave downwelling radiation, alongside the rainfall, sea-level pressure, and temperature (daily mean, min \& max) are included. To enable reproducible research and minicube generation anywhere on Earth, we open source a Python package called \emph{EarthNet Minicuber}\footnote{\url{https://github.com/earthnet2021/earthnet-minicuber}}, which generates multi-modal minicubes in a cloud native manner: only downloading the data chunks actually needed, instead of a full Sentinel 2 tile.

\begin{figure*}[t]
    \centering
    \includegraphics[width=\textwidth]{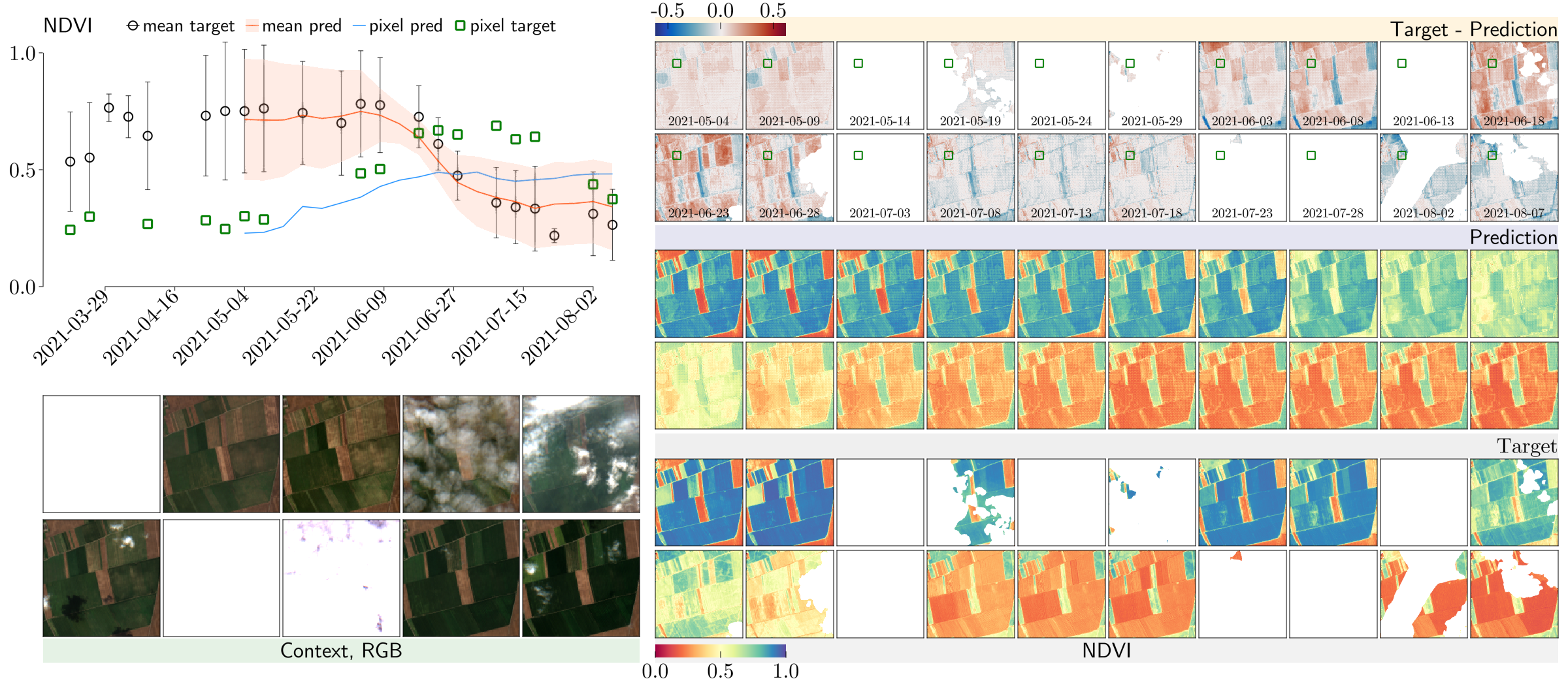}
    \caption{Qualitative Results of Contextformer for one OOD-t minicube located near Oradea, Romania. The top-left shows timeseries for all pixels (mean and std. dev.) and for a single pixel (green square on top right). The right side shows image timeseries of cloud-masked target and predicted NDVI alongside their difference.}
    \label{fig:qualitative}
\end{figure*}

\textbf{Cloud mask} Vegetation proxies derived from optical satellite imagery are only meaningful if observations with clouds, shadows and snow are excluded such that anomalies due to clouds can be distinguished from vegetation anomalies. We train a UNet with Mobilenetv2 encoder \cite{sandler.etal_2018} on the CloudSEN12 dataset \cite{aybar.etal_2022} to detect clouds and cloud shadows from RGB and Nir bands. Tab.~\ref{tab:cloudmask} compares precision, recall and F1 scores for detecting faulty pixels. Our approach outperforms Sen2Cor \cite{louis.etal_2016} (used in EarthNet2021), FMask \cite{qiu.etal_2019} and KappaMask \cite{domnich.etal_2021} baselines by a large margin. If using Sen2Cor in addition, to allow for snow masking, precision drops, but recall increases: i.e. the cloud mask gets more conservative. Using all 13 Sentinel 2 L2A bands is better than just using 4 bands, however such a model is not directly applicable to GreenEarthNet.

\textbf{Test sets} Due to meso-scale circulation patterns, weather has high spatial correlation lengths. For GreenEarthNet, we design test sets ensuring independence not only in the high-resolution satellite data but also in the coarse-scale meteorology between training and test minicubes.
More specifically, we introduce the subsets
\begin{itemize}[noitemsep,topsep=0pt]
    \item \emph{Train}, 23816 minicubes in years 2017-2020
    \item \emph{Val} 245 minicubes close to training locations, year 2020
    \item \emph{OOD-t test} same locations as Val, years 2021-2022
    \item \emph{OOD-s test}, 800 minicubes stratified over $1\degree \times 1\degree$ lat-lon grid cells outside training regions, years 2017-2019
    \item \emph{OOD-st test} same as OOD-s, but years 2021-2022
\end{itemize}
\emph{OOD-t} is the main test set used throughout this study. It tests the models' ability to extrapolate in time: i.e. we allow it to learn from past information about a location and want to know how it would perform in the future. \emph{val} follows the same reasoning and hence allows for early stopping of models according to their temporal extrapolation skill. \emph{OOD-s} and \emph{OOD-st} test spatial extrapolation, as well as spatio-temporal extrapolation. The test sets are compatible with EarthNet2021: the \emph{Val}/\emph{OOD-t} locations are inside the EarthNet2021 IID tests set and the \emph{OOD-s}/\emph{OOD-st} locations are far away from EarthNet2021 training data. For all test sets, we create minicubes over four periods during the European growing season \cite{rotzer.chmielewski_2001} each year: Predicting March-May (MAM), May-July (MJJ), July-September (JAS) and September-November (SON).

\textbf{Additional Layers} We add the ESA Worldcover Landcover map \cite{zanagadaniele.etal_2021} for selecting only vegetated pixels during evaluation, the Geomorpho90m Geomorphons map \cite{amatulli.etal_2020} for further evaluation and the ALOS \cite{tadono.etal_2016}, Copernicus \cite{esa_2021} and NASA \cite{crippen.etal_2016} DEMs, to provide uncertainty in the elevation maps. Finally, we provide georeferencing for each minicube, enabling their extension with further data.

\subsection{Evaluation}
We resort to traditional metrics in environmental modeling:
\begin{itemize}[noitemsep,topsep=0pt]
    \item $R^2$, the squared pearson correlation coefficient
    \item $\text{RMSE}$, the root mean squared error
    \item $\text{NSE} = 1 - \dfrac{MSE(V, \hat{V})}{Var[V]}$, the nash-sutcliffe efficiency \cite{nash.sutcliffe_1970}, a measure of relative variability
    \item $|\text{bias}| = | \overline{V} - \overline{\hat{V}} |$, the absolute bias
\end{itemize}
In addition, we propose to measure if a model is better than the NDVI climatology, by computing the \emph{Outperformance score}: the percentage of minicubes, for which the model is better in at least 3 out of the 4 metrics. Here, better means their score difference (ordering s.t. higher=better) exceeds $0.01$ for RMSE and $|\text{bias}|$ and $0.05$ for NSE and $R^2$. We also report the RMSE over only the first 25 days (5 time steps) of the target period. 

We compute all metrics per pixel over clear-sky timesteps. We then consider only pixels with vegetated landcover (cropland, grassland, forest, shrubland), no seasonal flooding (minimum NDVI $>0$), enough observations ($\geq10$ during target period, $\geq3$ during context period) and considerable variation (NDVI std. dev $>0.1$). All these pixelwise scores are grouped by minicube and landcover, and then aggregated to account for class imbalance. Finally, the macro-average of the scores per landcover class is computed. In this way, the scores represent a conservative estimate of the expected performance of dynamic vegetation modeling during a new year or at a new location.

\begin{table*}[t]
    \centering
    \begin{tabularx}{\textwidth}{>{\footnotesize\scshape}cXccccccc}
    \toprule
    & Model & $R^2$ $\uparrow$ & RMSE $\downarrow$ & NSE $\uparrow$ &   $|\text{bias}|$ $\downarrow$ & $\underset{\text{Climatology}}{\text{Outperform}}$ $\uparrow$& $\underset{25 \text{ days}}{\text{RMSE}}$ $\downarrow$& \#Params\\
    \midrule
    \parbox[t]{1mm}{\multirow{3}{*}{\rotatebox[origin=c]{90}{non-ML}}}& Persistence & 0.00 &      0.23 &      -1.28 &      0.17 &      21.8\% &      0.09 & 0 \\
    & Previous year &  0.56 &      0.20 &      -0.40 &      0.14 &      19.3\% &      0.18  & 0 \\
    & Climatology & 0.58 &      0.18 &      -0.34 &      0.13 &       n.a. &      0.16 & 0 \\
    \arrayrulecolor{black!30}\midrule
    \parbox[t]{1mm}{\multirow{3}{*}{\rotatebox[origin=c]{90}{local TS}}} & Kalman filter &      0.41 &      0.19 &      -0.57 &      0.13 &      27.0\% &      0.16  & $\mathcal{O}$(10)\\
    & LightGBM &      0.51 &      0.17 &      -0.22 &      0.12 &      42.2\% &      0.11  & n.a.\\
    & Prophet & 0.57 &      0.16 &      -0.05 &      0.11 &      60.6\% &      0.13  & $\mathcal{O}$(10) \\
    \arrayrulecolor{black!30}\midrule
    \parbox[t]{1mm}{\multirow{3}{*}{\rotatebox[origin=c]{90}{EN21}}} & ConvLSTM \cite{diaconu.etal_2022} & 0.51 &      0.18 &      -0.37 &      0.12 &      43.9\% &      0.12  & 0.2M \\
    & SG-ConvLSTM \cite{kladny2024enhanced} & 0.53 &      0.19 &      -0.33 &      0.14 &      45.8\% &      0.11  & 0.7M \\
    & Earthformer \cite{gao.etal_2022} &  0.49 &      0.17 &      -0.27 &      0.12 &      47.2\% &      0.11  &  60.6M\\
    \arrayrulecolor{black!30}\midrule
    \parbox[t]{1mm}{\multirow{5}{*}{\rotatebox[origin=c]{90}{This Study}}} & ConvLSTM \cite{robin.etal_2022} & 0.58 \footnotesize{$\pm$0.01} & 0.16 \footnotesize{$\pm$0.00} &  -0.13 \footnotesize{$\pm$0.02} & 0.11 \footnotesize{$\pm$0.00} & 53.1\% \footnotesize{$\pm$1.2\%} & 0.11 \footnotesize{$\pm$0.00} & 1.0M \\
    & Earthformer \cite{gao.etal_2022} & 0.52 &      0.16 &      -0.13 &      0.10 &      56.5\% & 0.09 & 60.6M\\
    & PredRNN \cite{wang.etal_2023} & \textbf{0.62} \footnotesize{$\pm$0.00} & 0.15 \footnotesize{$\pm$0.00} &  0.03 \footnotesize{$\pm$0.00} & 0.10 \footnotesize{$\pm$0.00} & 64.7\% \footnotesize{$\pm$1.2\%} & 0.10 \footnotesize{$\pm$0.00} & 1.4M \\
    & SimVP \cite{tan.etal_2023} & 0.60 \footnotesize{$\pm$0.00} & 0.15 \footnotesize{$\pm$0.00} &  0.03 \footnotesize{$\pm$0.01} & \textbf{0.09} \footnotesize{$\pm$0.00} & 64.1\% \footnotesize{$\pm$1.0\%} & 0.10 \footnotesize{$\pm$0.00} & 6.6M \\
    & Contextformer (Ours) & \textbf{0.62} \footnotesize{$\pm$0.00} & \textbf{0.14} \footnotesize{$\pm$0.00} &  \textbf{0.09} \footnotesize{$\pm$0.01} & \textbf{0.09} \footnotesize{$\pm$0.00} & \textbf{66.8\%} \footnotesize{$\pm$0.3\%} & \textbf{0.08} \footnotesize{$\pm$0.00} & 6.1M \\
    \arrayrulecolor{black}\bottomrule
    \end{tabularx}
    \caption{Quantitative Results. Mean ($\pm$std. dev.) are computed from three different random seeds.}
    \label{tab:quantitative}
\end{table*}

\subsection{Baselines}\label{sec:baselines}
We evaluate Contextformer against diverse baselines representative of various model classes, including non-ML methods, time series forecasting models, the top 3 performers on the EarthNet2021 benchmark \cite{requena-mesa.etal_2021}, classical, and two state-of-the-art video prediction models. This choice aims to account for uncertainty regarding the optimal models for vegetation forecasting, considering factors such as the relevance of spatial context. While existing spatio-temporal earth surface forecasting models are expected to serve as strong baselines due to task similarity, recent advancements in video prediction, leveraging the perspective of satellite image time series as a video, may also offer a competitive advantage.




\textbf{Non-ML baselines} We evaluate three non-ML baselines related to ecological memory: persistence \cite{requena-mesa.etal_2021} (last cloud free NDVI pixel), previous year \cite{robin.etal_2022} (linearly interpolated) and climatology (mean NDVI seasonal cycle). 

\textbf{Local time series models} We compare against three common time series models: Kalman filter, LightGBM \cite{ke.etal_2017} and Prophet \cite{taylor.letham_2018} from the Python library darts \cite{herzen.etal_2022}. These are trained on timeseries from a single pixel and applied to forecast this pixel, given future weather as covariates. They are expensive to run: a single minicube takes $\sim3$h on an 8-CPU machine, $\mathcal{O}(10^4)$ slower than deep learning. We also evaluate a global timeseries model: the LSTM (implemented as ConvLSTM with 1x1 kernel). The time series models should be strong if spatial interactions are less predictive for vegetation evolution.

\textbf{EarthNet2021 models} 
We also evaluate the Top-3 models from the EarthNet2021 challenge leaderboard\footnote{\url{https://web.archive.org/web/20230228215255/https://www.earthnet.tech/en21/ch-leaderboard/}} using their trained weights: a regular ConvLSTM \cite{diaconu.etal_2022}, an encode-process-decode ConvLSTM called SGED-ConvLSTM \cite{kladny2024enhanced} and the Earthformer \cite{gao.etal_2022}, a transformer model using cuboid-attention. 

Additionally, we train and fine-tune both the ConvLSTM and Earthformer on GreenEarthNet. For the ConvLSTM, we follow the original Shi et al. \cite{shi.etal_2015} encoding-forecasting setup, which is different from ConvLSTM flavors studied on EarthNet2021 \cite{diaconu.etal_2022, kladny2024enhanced} but has demonstrated improved performance on a similar problem in Africa \cite{robin.etal_2022}. We condition the Earthformer \cite{gao.etal_2022} through early fusion during historical steps and latent fusion during future steps.


\textbf{Video prediction models}
We adapt two state-of-the-art video prediction models (PredRNN and SimVP) and two basic UNet-based approaches. The next-frame UNet \cite{rasp.etal_2020} predicts autoregressively one step ahead. The next-cuboid UNet \cite{requena-mesa.etal_2021} directly predicts all time steps, taking the historical time steps stacked in the channel dimension. PredRNN \cite{wang.etal_2017, wang.etal_2023} is an autoregressive model with improved information flow. We generalize the action-conditioned PredRNN \cite{wang.etal_2023} by using feature-wise linear modulation \cite{perez.etal_2018} for weather conditioning on the inputs. SimVP \cite{tan.etal_2023} performs direct multi-step prediction through an encode-process-decode ConvNet, we adapt it with weather conditioning by feature-wise linear modulation \cite{perez.etal_2018} on the latent embeddings at each stage of the processor.

\subsection{Implementation details} We build all of our ConvNets with a PatchMerge-style architecture similar to the one in Earthformer \cite{gao.etal_2022}. For SimVP and PredRNN, such encoders and decoders are more powerful, but also slightly more parameter-intensive, than the variants used in the original papers. We use GroupNorm \cite{wu.he_2018} and LeakyReLU activation \cite{xu.etal_2015b} for the ConvNets, and  and ConvLSTMs. For the Contextformer, we use LayerNorm \cite{ba2016layer} and GELU activation \cite{hendrycks2023gaussian}. For ConvNets, skip connections preserve high-fidelity content between encoders and decoders. Our framework is implemented in PyTorch, and models are trained on Nvidia A40 and A100 GPUs. We use the AdamW \cite{loshchilov.hutter_2022} optimizer and tune the learning rate and a few hyperparameters per model. More implementation details can be found in the supplementary materials.

\section{Experiments}

\subsection{Baseline comparison}
We conduct experiments for predicting vegetation state across Europe in 2021 and 2022 at $20m$ resolution and compare the Contextformer against a wide range of baselines. The quantitative results are shown in table~\ref{tab:quantitative}. For Contextformer, ConvLSTM, PredRNN and SimVP, we report the mean ($\pm$std. dev.) from three different random seeds. Earthformer has an order of magnitude more parameters, making training more expensive, which is why we only report one random seed. We find the Contextformer outperforms (or performs on par with) every baseline on all metrics. It achieves $R^2 = 0.62$ and $0.14$ RMSE on the full 100 days lead time, which is further improved to $0.08$ RMSE during the first 25 days lead time. The closest competitors are PredRNN and SimVP, with PredRNN having on par $R^2 = 0.62$ and SimVP on par $|\text{bias}| = 0.09$.

The Contextformer and the other video prediction baselines trained in this study are the first models to outperform the Climatology baseline: the ConvLSTM reaches $53.1\%$ outperformance score, while the Contextformer achieves $66.8\%$ (with consistent ranking across thresholds, see sec.~\ref{sec:robustoutpf}). For the top-3 models (PredRNN, SimVP and Contextformer) and all metrics, differences to the climatology are highly significant when tested for all pixels (with Wilcoxon signed-rank test, $\alpha = 0.001$), but also for each land cover or for smaller subsets of $100$ minicubes. ConvLSTM and Earthformer have overall lower skill. They mostly excels at RMSE and $|\text{bias}|$, where they can perform similar to other methods, yet have way lower performance for NSE and $R^2$.

\begin{figure}
    \centering
    \includegraphics[width=\columnwidth]{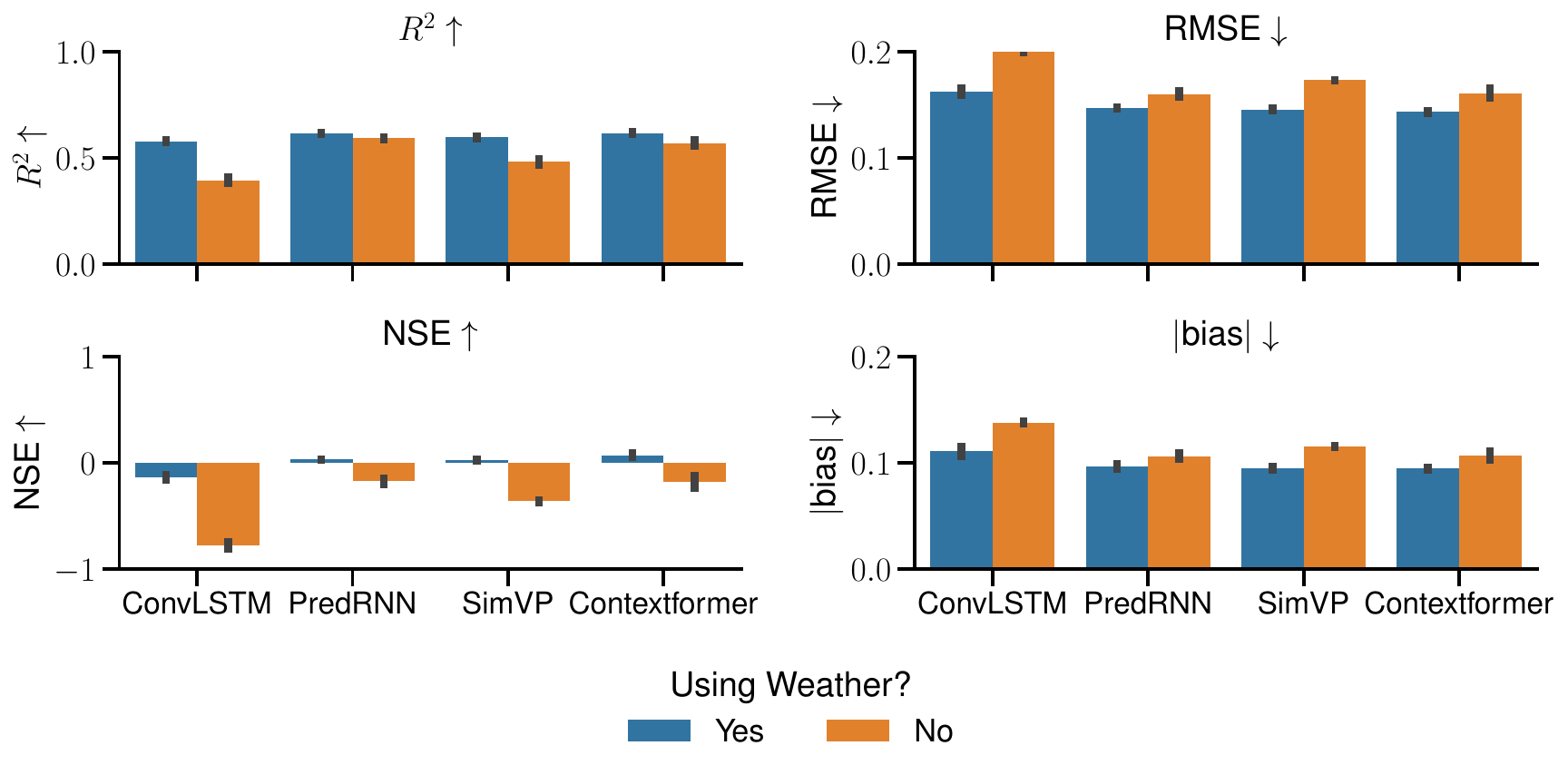}
    \caption{Model performance comparing meteo-guided models (blue) with the ablation not using weather (black bar is std. dev. from three random seeds).}
    \label{fig:horizon}
\end{figure}

The models trained on EarthNet2021 (ConvLSTM \cite{diaconu.etal_2022}, SGED-ConvLSTM \cite{kladny2024enhanced} and Earthformer \cite{gao.etal_2022}) perform poorly. None of the approaches consistently beats the Climatology, particularly the $R^2$ is much lower (from $0.49$ for Earthformer up to $0.53$ for SGED-ConvLSTM). Likely, this is a result of the focus on perceptual quality that was reflected in the EarthNet2021 metrics, as well as the overall lower data quality due to a faulty cloud mask. 

Finally, other local time series baselines and non-ML baselines also underperform the Contextformer. The strongest pixelwise model is Prophet \cite{taylor.letham_2018}, with an outperformance score of $60.6\%$, followed by the climatology. Note, all of these baselines have access to a lot more information than the deep learning-based models (6 years vs. 50 days). Hence, this model comparison gives an indication, that spatial context is useful for vegetation forecasting, but leveraging them is challenging, as temporal dynamics are more dominant. In addition, the local time series models are all very slow, compared to the deep learning solutions presented in this work, which perform predictions within seconds (see sec.~\ref{sec:speed}).

\begin{table}
\centering
\begin{tabularx}{\columnwidth}{Xccc}
\toprule
& Original & Shuffled & \\
Model & $R^2$ $\uparrow$ & $R^2$ $\uparrow$ & Diff $\uparrow$ \\
\midrule
Climatology & 0.58 & - & - \\
\arrayrulecolor{black!30}\midrule
1x1 LSTM & 0.53 & 0.53 & 0.00 \\
Next-frame UNet & 0.51 & 0.48 & -0.03 \\
Next-cuboid UNet & 0.56 & 0.43 & -0.13 \\
\arrayrulecolor{black!30}\midrule
ConvLSTM & 0.58 & 0.46 & -0.12 \\
PredRNN & 0.62 & 0.45 & -0.17 \\
SimVP & 0.60 &  0.49 & -0.11 \\
Contextformer & 0.62 & 0.55 & -0.07 \\
\arrayrulecolor{black}\bottomrule
\end{tabularx}
\caption{Model skill when spatial interactions are broken through shuffling.}
\label{tab:spatiotemp}
\end{table}

Qualitative results of the Contextformer model for one minicube are reported in fig.~\ref{fig:qualitative}. The model clearly learns the complex dynamics of vegetation, with a strong seasonal evolution of the crop fields. It interpolates faithfully those pixels, which are masked in the target, and contains strong temporal consistency. However, as the lead time increases, predictions become less explicit, with a tendency towards oversmoothing.

\subsection{Weather guidance}
Our meteo-guided models benefit from the weather conditioning. Fig.~\ref{fig:horizon} compares ConvLSTM, PredRNN, SimVP and Contextformer (blue) against a variant without weather conditioning (orange). For all metrics, using weather outperforms not using it. The ConvLSTM has the largest performance gain due to meteo-guidance, yet it is also the weakest model. This could possibly be due to the ConvLSTMs smaller receptive field and hence lower capacity at leveraging spatial context, which may to some degree compensate predictive capacity from weather. 

For PredRNN and SimVP, we conduct an extended ablation study on weather guidance (see supplementary material). Weather conditioning methods (concatenation, FiLM \cite{perez.etal_2018}, and cross-attention \cite{rombach.etal_2022}) have a minor impact on performance when applied appropriately: cross-attention is most useful with latent fusion, FiLM outperforms concatenation, and is suitable for early fusion.

\subsection{The role of spatial interactions}
Unlike video prediction, satellite images show minimal spatial movement. Field and forest boundaries remain mostly fixed, with the largest variations occurring within these edges over time. Hence, it is unclear whether spatio-temporal models, accounting for interactions, are suitable for modeling vegetation dynamics. However, at $20m$ resolution, lateral processes may occur, not captured by predictors. For example, grasslands near a river or on a north-facing slope may react differently to meteorological drought. Additionally, weather affects trees at the forest edge differently from those in the center.

We compare model performance with spatially shuffled input, i.e. explicitly breaking spatial interactions \cite{requena-mesa.etal_2019}. We shuffle across batch and space, to also destroy image statistics. We evaluate Contextformer, ConvLSTM, PredRNN, and SimVP, skipping Earthformer due to high training cost. In addition we also study three baselines: a pixelwise (1x1) LSTM, the next-frame UNet and the next-cuboid UNet (see sec.~\ref{sec:baselines}). The pixelwise LSTM is a global timeseries model unable to capture spatial interactions. The next-frame UNet models spatial interactions, but does not consider temporal memory. All other models can leverage spatio-temporal dependencies, though the ConvLSTM only has a small local receptive field ($\sim100$m around each pixel). The results are reported in tab.~\ref{tab:spatiotemp}. As can be expected, the pixelwise LSTM can be trained with spatial shuffled pixels without performance loss. All other models, though, exhibit a drop in performance under pixel shuffling. For Contextformer, ConvLSTM, PredRNN and SimVP, $R^2$ drops by at least $0.07$ and $RMSE$ increases by at least $0.04$. 

\begin{table}
\centering
\begin{tabularx}{\columnwidth}{Xcccc}
\toprule
Ablation & $R^2$ $\uparrow$ & RMSE $\downarrow$ & $\underset{\text{Climatology}}{\text{Outperf}}$ $\uparrow$ \\
\midrule
MLP vision encoder & 0.58 & 0.15 & 58.3\%\\
\arrayrulecolor{black!30}\midrule
PVT encoder (frozen) & 0.57 & 0.17 & 46.1\% \\
PVT encoder & 0.62 & 0.15 & 62.3\% \\
\arrayrulecolor{black!30}\midrule
\quad /w cloud mask token & 0.61 & 0.16 & 61.8\% \\
\quad /w learned $\hat{V}^0$ & 0.62 & 0.16 & 60.6\% \\
\quad /w last pixel $\hat{V}^0$ & 0.62 & 0.15 & 65.1\% \\
\arrayrulecolor{black!30}\midrule
Contextformer-6M & 0.62 & 0.14 & 66.8\% \\
Contextformer-16M & 0.61 & 0.14 & 67.3\% \\
\arrayrulecolor{black}\bottomrule
\end{tabularx}
\caption{Model ablations. The Contextformer uses a PVT encoder, a cloud mask token and the last cloud free pixel as $\hat{V}_0$.}
\label{tab:ablation}
\end{table}

\subsection{Ablation Study of Contextformer components}
We conduct experiments to show how each key component in our Contextformer affects predictive skill. Tab.~\ref{tab:ablation} lists the results of our ablation studies. First, we find that continued training of a pre-trained PVT vision encoder (outperformance score $62.3$\%) outperforms both a MLP vision encoder and a frozen pre-trained PVT. Second, adding the delta-prediction scheme with an initial vegetation state estimate $\hat{V}^0$ constructed by the last historical cloud free NDVI pixel further improves the outperformance to $65.1\%$ -- the version directly predicting NDVI is \emph{PVT encoder}. Instead using a learned MLP decoder to estimate $\hat{V}^0$ is inferior. Third, using the cloud mask to drop out faulty tokens from the PVT encoder decreases model skill, if used alone, but if used on top of the delta-prediction scheme in the final model Contextformer-6M, it gives another boost to $66.8\%$ outperformance. Finally, scaling the model size of the Contextformer to 16M parameters is not helpful when trained on GreenEarthNet, indicating the need for an even larger dataset for further performance gains.

\begin{figure}
    \centering
    \includegraphics[width=\columnwidth]{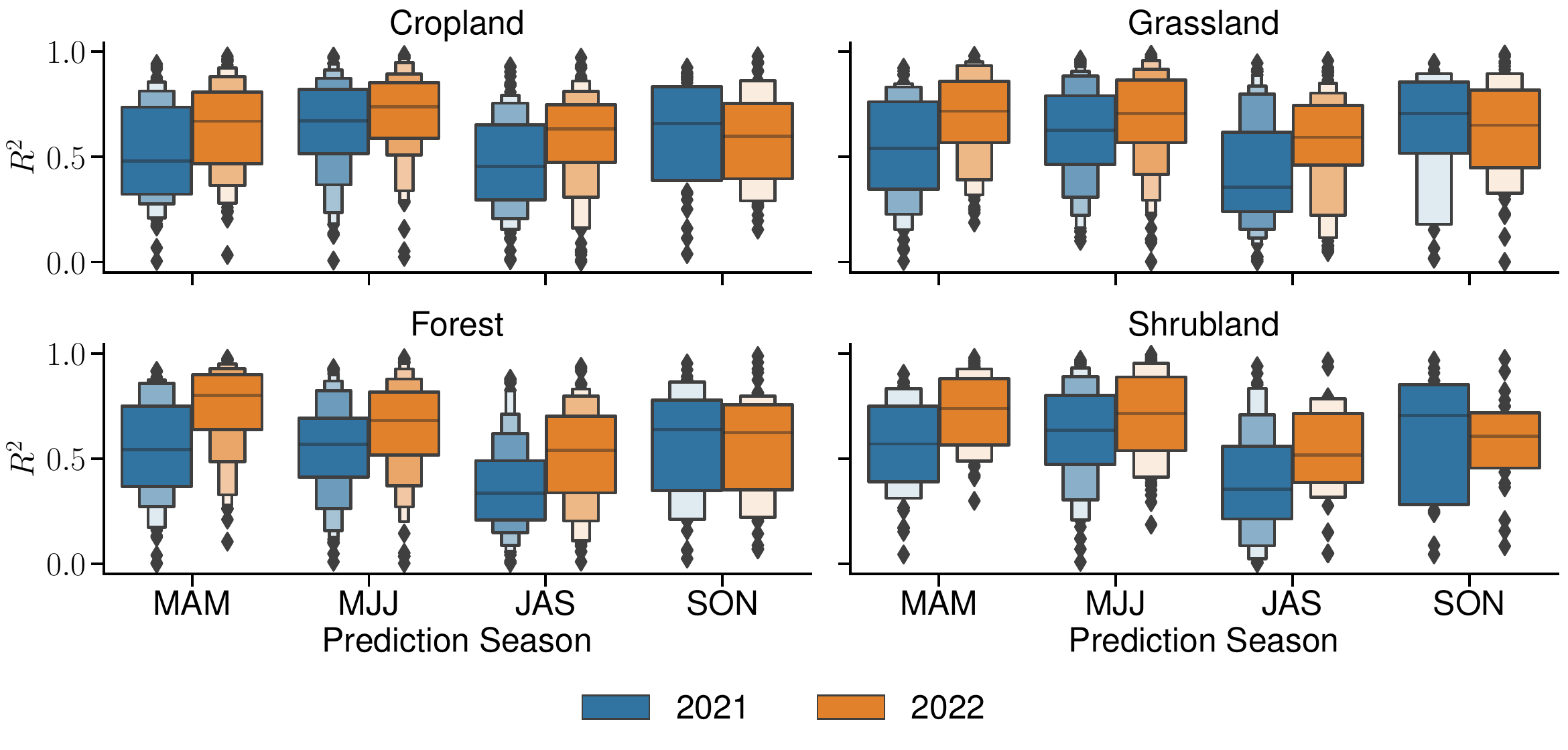}
    \caption{Contextformer model skill for different seasons and landcover on the OOD-t test set.}
    \label{fig:seasons}
\end{figure}

\begin{table}
\centering
\begin{tabularx}{\columnwidth}{Xcccc}
\toprule
& \multicolumn{2}{c}{OOD-s} & \multicolumn{2}{c}{OOD-st} \\
Model & $R^2$ $\uparrow$ & RMSE $\downarrow$ & $R^2$ $\uparrow$ & RMSE $\downarrow$ \\
\midrule
Climatology & 0.50 & 0.15 & 0.56 & 0.19 \\
Contextformer & 0.54 & 0.15 & 0.58 & 0.14 \\
\bottomrule
\end{tabularx}
\caption{Model skill at spatial (OOD-s) and spatio-temporal (OOD-st) extrapolation.}
\label{tab:extrapol}
\end{table}

\subsection{Contextformer Strengths and Limitations}

The OOD-t test set includes minicubes from four 3-month periods over two years. Fig.~\ref{fig:seasons} shows Contextformer's model skill. Yearly variations are significant. Growing season prediction was better in 2022 until September, then it switched, and 2021 performed better. First half of the season is usually better predicted than the second half, likely due to anthropogenic influences (harvest, mowing, cutting, and forest fires). These events are challenging to predict from weather covariates and may be interpreted as random noise.

We assess the performance at spatio-(temporal) extrapolation of the Contextformer on the OOD-s and OOD-st test sets and report in tab.~\ref{tab:extrapol}. The Contextformer can extrapolate in space and time. However, the margin to the climatology does shrink. Here, more training data might help: spatial extrapolation is theoretically not necessary for modeling vegetation dynamics (only temporal extrapolation is). Practically speaking, however, it does help to increase inference speed and enable potential applicability over large areas.

For these practical situations another aspect needs to be studied in future work: at inference time weather comes from uncertain weather forecasts. Here, we first wanted to learn the impact of weather on vegetation and thus took the historical meteo data which has the least error. We expect the weather forecast uncertainty (represented by realizations / scenarios) to mostly propagate, but not present a covariate shift larger than the inter-annual variability, which our models are robust to (OOD-t evaluation).

\section{Conclusion}
We proposed Contextformer, a multi-modal transformer model designed for fine-resolution vegetation greenness forecasting. It leverages spatial context through a Pyramid Vision Transformer backbone while maintaining parameter efficiency. The temporal component is a transformer that independently models the dynamics of local context patches over time, incorporating meteorological data. We additionally introduce the novel GreenEarthNet dataset tailored for self-supervised vegetation forecasting and compare Contextformer against an extensive set of baselines.

Contextformer outperforms the previous state-of-the-art, especially on nash-sutcliffe efficiency and surpasses even strong freshly trained video prediction baselines. To our knowledge, we are the first to consider a climatology baseline and outperforming it with models. Given the pronounced seasonality of vegetation dynamics, this suggests real-world applicability for our models, particularly the Contextformer, in crucial scenarios like humanitarian anticipatory action or carbon monitoring.


{
\paragraph{Code.} We provide code and pre-trained weights to reproduce our experimental results under \url{https://github.com/vitusbenson/greenearthnet} \cite{benson_2024_10793870}.\\
\textbf{Author contributions.} VB experiments, figures, writing. CR experiments, writing. CRM supervision, figures, writing. ZG experiments. LA figures. NC, JC, NL, MW writing. MR funding, supervision, writing. All authors contributed to discussing the results. \\
\textbf{Acknowledgments.} We are thankful for invaluable help, comments and discussions to Reda ElGhawi, Christian Reimers, Annu Panwar and Xingjian Shi. MW thanks the European Space Agency for funding the project DeepExtremes (AI4Science ITT). CRM and LA are thankful to the European Union’s DeepCube Horizon 2020 (research and innovation programme grant agreement No 101004188). NL and JC acknowledge funding from the European Union’s Horizon 2020 research and innovation programme under grant agreement No 101003469. 
}
\clearpage
{\small
\bibliographystyle{ieeenat_fullname}
\bibliography{earthnet}
}

\clearpage
\appendix
\section{Model details}
\subsection{Cloud masking}
\paragraph{Baselines (Table 1)}
The baselines reported in table 1 are taken from CloudSEN12 \cite{aybar.etal_2022}. Sen2Cor \cite{louis.etal_2016} is the processing software from ESA used to produce the Scene Classification Layer (SCL) mask, which was also introduced in EarthNet2021 \cite{requena-mesa.etal_2021}. FMask \cite{qiu.etal_2019} is a processing software originally designed for NASA Landsat imagery, but now repurposed to also work with Sentinel 2 imagery. It requires L1C top-of-atmosphere reflectance from all bands to be produced (EarthNet2021 only containes L2A bottom-of-atmosphere reflectance from four bands). KappaMask \cite{domnich.etal_2021} is a cloud mask based on deep learning, in table 1 we reported scores from the L2A version, which uses all 13 L2A bands as input.

\paragraph{UNet Mobilenetv2 (Table 1)}
Our UNet with Mobilenetv2 encoder \cite{sandler.etal_2018} was trained in two variants, one with RGB and near-infrared bands of L2A imagery (i.e. works with EarthNet2021) and one with all 13 bands of L2A imagery. We adopted the exact same implementation that was benchmarked in the CloudSEN12 paper \cite{aybar.etal_2022}, with the only difference being that in the paper, L1C imagery was used (which is often not useful in practical use-cases). In detail, this means we trained the UNet with Mobilenetv2 encoder using the Segmentation Models PyTorch Python library\footnote{\url{https://segmentation-models-pytorch.readthedocs.io/en/latest/}}. We used a batch size of 32, random horizontal and vertical flipping, random 90 degree rotations, random mirroring, unweighted cross entropy loss, early stopping with a patience of 10 epochs, AdamW optimizer, learning rate of $1e^{-3}$, and a learning rate schedule reducing the learning rate by a factor of 10 if validation loss did not decrease for 4 epochs.

\subsection{Vegetation modeling}

\paragraph{Contextformer (Table 2,3,4,5 Figure 3,4,5)}
Our Contextformer is a combination of a spatial vision encoder: Pyramid Vision Transformer (PVT) v2 B0 \cite{wang.etal_2021a, wang.etal_2022c} with pre-trained ImageNet1k weights from the PyTorch Image Models library\footnote{\url{https://github.com/rwightman/pytorch-image-models}} and a temporal transformer encoder. We use a patch size of $4\times 4$ px. We use an embedding size of $256$ and the temporal transformer has three self-attention layers with $8$ heads, each followed by an MLP with $1024$ hidden channels. We use LayerNorm \cite{ba2016layer} for normalization and GELU \cite{hendrycks2023gaussian} as non-linear activation function. The model is trained with masked token modeling, randomly ($p=0.5$) flipping between inference mask (future token masked) and random dropout mask ($70\%$ of image patches masked, except for the first 3 time steps). We train for 100 epochs with a batch size of 32, a learning rate of $4e^{-5}$ and with AdamW optimizer on $2$ NVIDIA A100 GPUs. We train three models from the random seeds 42, 97 and 27.

\paragraph{Local timeseries models (Table 2)}
We train the local timeseries models (table 2) at each pixel. For a given pixel we extract the full timeseries of NDVI and weather variables at 5-daily resolution. All variables are linearly gapfilled and weather is aggregated with min, mean, max, and std to 5-daily. The whole timeseries before each target period is used to train a timeseries model, for the target period the model only receives weather. The Kalman Filter runs with default parameters from darts \cite{herzen.etal_2022}. The LightGBM model gets lagged variables from the last 10 time steps and predicts a full 20 time step chunk at once. For Prophet we again use default parameters.

\paragraph{EarthNet models (Table 2)}
For running the leading models from EarthNet2021 we utilize the code from the respective github repositories: ConvLSTM \cite{diaconu.etal_2022}\footnote{\url{https://github.com/dcodrut/weather2land}}, SGED-ConvLSTM \cite{kladny2024enhanced}\footnote{\url{https://github.com/rudolfwilliam/satellite_image_forecasting}} and Earthformer \cite{gao.etal_2022} \footnote{\url{https://github.com/amazon-science/earth-forecasting-transformer/tree/main/scripts/cuboid_transformer/earthnet_w_meso}}. We derive the NDVI from the predicted satellite bands red and near-infrared:
\begin{align}
    NDVI = \frac{NIR - Red}{NIR + Red + 1e^{-8}}
\end{align}

\paragraph{ConvLSTM (Table 2,3, Figure 4,5)}
Our ConvLSTM contains four ConvLSTM-cells \cite{shi.etal_2015} in total, two for processing context frames and two for processing target frames. Each has convolution kernels with bias, hidden dimension of 64 and kernel size of 3. We train for 100 epochs with a batch size of 32, a learning rate of $4e^{-5}$ and with AdamW optimizer. We train three models from the random seeds 42, 97 and 27.

\paragraph{PredRNN (Table 2,3, Figure 4)}
Our PredRNN contains two ST-ConvLSTM-cells \cite{wang.etal_2017} Each has convolution kernels with bias, hidden dimension of 64 and kernel size of 3 and residual connections. We use a PatchMerge encoder decoder with GroupNorm (16 groups), convolutions with kernel size of 3 and hidden dimension of 64, LeakyReLU activation and downsampling rate of 4x. We train for 100 epochs with a batch size of 32, a learning rate of $3e^{-4}$ and with AdamW optimizer. We use a spatio-temporal memory decoupling loss term with weight 0.1 and reverse exponential scheduling of true vs. predicted images (as in the PredRNN journal version \cite{wang.etal_2023}). We train three models from the random seeds 42, 97 and 27.

\paragraph{SimVP (Table 2,3, Figure 4)}
Our SimVP has a PatchMerge encoder decoder with GroupNorm (16 groups), convolutions with kernel size of 3 and hidden dimension of 64, LeakyReLU activation and downsampling rate of 4x. The encoder processes all 10 context time steps at once (stacked along the channel dimension). The decoder processes 1 target time step at a time. The gated spatio-temporal attention processor \cite{tan.etal_2023} translates between both in the latent space, we use two layers and 64 hidden channels. We train for 100 epochs with a batch size of 64, a learning rate of $6e^{-4}$ and with AdamW optimizer. We train three models from the random seeds 42, 97 and 27.

\paragraph{Earthformer (Table 2)}
Our Earthformer is a transformer combined with an initial PatchMerge encoder (and a final decoder) to reduce the dimensionality. The encoder and decoder use LeakyReLU activation, hidden size of 64 and 256 and downsample 2x. In between, the transformer processor has a UNet-type architecture, with cross-attention to merge context frame information with target frame embeddings. GeLU activation and LayerNorm, axial self-attention, 0.1 dropout and 4 attention heads are used. Weather information is regridded to match the spatial resolution of satellite imagery and used as input during context and target period. We train for 100 epochs with a batch size of 32, a maximum learning rate of $1e^{-4}$, linear learning rate warm up, cosine learning rate shedule and with AdamW optimizer.

\paragraph{1x1 LSTM (Table 4)}
Our 1x1 LSTM is implemented as a ConvLSTM with kernel size of 1. We train for 100 epochs with a batch size of 32, a learning rate of $4e^{-5}$ and with AdamW optimizer.

\paragraph{Next-frame UNet (Table 4)}
Our next-frame UNet has a depth of 5, latent weather conditioning with FiLM, a hidden size 128, kernel size 3, LeakyReLU activation, GroupNorm (16 groups), PatchMerge downsampling and nearest upsampling. We train for 100 epochs with a batch size of 64, a learning rate of $6e^{-4}$ and with AdamW optimizer.

\paragraph{Next-cuboid UNet (Table 4)}
Our next-cuboid UNet has a depth of 5, latent weather conditioning with FiLM, a hidden size 256, kernel size 3, LeakyReLU activation, GroupNorm (16 groups), PatchMerge downsampling and nearest upsampling. We train for 100 epochs with a batch size of 64, a learning rate of $6e^{-4}$ and with AdamW optimizer.

\section{Weather ablations}
\subsection{Methods}
Most of our baseline approaches have been originally proposed to handle only past covariates. Here, we condition forecasts on future weather. A-priori it is not known how to best achieve this weather conditioning. For PredRNN and SimVP, we compare three approaches, each fused at three different locations. The approaches operate pixelwise, taking features $x_{in} \in \mathbb{R}^d$ and conditioning input $c_{i} \in \mathbb{R}^{n}$ for weather variable $i$. The conditioning layers $g(\cdot, \cdot; \phi)$ with parameters $\phi$ then operate as
\begin{align}
    x_{out} = g(x_{in}, c; \phi) \in \mathbb{R}^d
\end{align}
We parameterize $g$ with neural networks. 

\paragraph{CAT} First concatenates $x_{in}$ and a flattened $c$ along the channel dimension, and then performs a linear projection to obtain $x_{out}$ of same dimensionality as $x_{in}$. In practice we implement this with a 1x1 Conv layer.

\paragraph{FiLM} Feature-wise linear modulation \cite{perez.etal_2018} generalizes the concatenation layer before. It produces $x_{out}$ with linear modulation:
\begin{align}
  x_{out} = x_{in} + \sigma(\gamma(c; \phi_{\gamma})\odot N(f(x_{in}; \phi_{f})) + \beta(c; \phi_{\beta}))
\end{align}
Here, $f$ is a linear layer, $\gamma$ and $\beta$ are MLPs, $N$ is a normalization layer and $\sigma$ is a pointwise non-linear activation function.

\paragraph{xAttn} Cross-attention is an operation commonly found in the Transformers architecture. In recent works on image generation with diffusion models it is used to condition the generative process on a text embedding \cite{rombach.etal_2022}. Inspired from this, we propose a pixelwise conditioning layer based on multi-head cross-attention. The input $x_{in}$ is treated as a single token query $Q$. Each weather variable $c_{i}$ is treated as individual tokens, from which we derive keys $K$ and values $V$. The result is then just regular multi-head attention $MHA$ in a residual block:
\begin{align}
    x_{out} &= x_{in} \\
    &+ f(N(MHA(Q(x_{in}; \phi_{Q}), K(c; \phi_{K}), V(c; \phi_{V}))); \phi_{f})
\end{align}
Here, $f$ is either a linear projection or a MLP and $N$ is a normalization layer.

Each of the three approaches we apply at three locations throughout the network:

\paragraph{Early fusion} Just fusing all data modalities before passing it to a model. This Early CAT has been previously used for weather conditioning in satellite imagery forecasting 

\paragraph{Latent fusion} In the encode-process-decode framework, encoders are meant to capture spatial, and not temporal, relationships. Hence, latent fusion conditions the encoded spatial inputs twice: right after leaving the encoder and before entering the decoder.

\paragraph{All (fusion everywhere)} In addition, we compare against conditioning at every stage of the encoders, processors and decoders. All CAT has been applied to condition stochastic video predictions on random latent codes \cite{lee.etal_2018}.

\subsection{Results}
\begin{figure}
    \centering
    \includegraphics[width=\columnwidth]{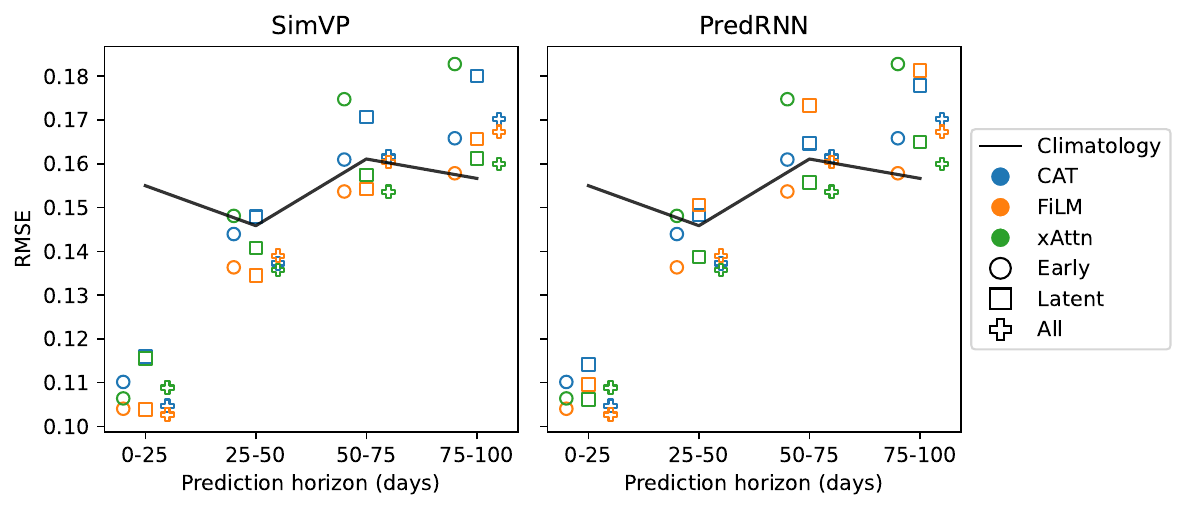}
    \caption{Model performance (RMSE) when using different ways of weather conditioning over varying prediction horizons.}
    \label{fig:horizon2}
\end{figure}

Fig.~\ref{fig:horizon2} summarizes the findings by looking at the RMSE over the prediction horizon. For the first 50 days, most models are better than the climatology, afterwards, most are worse. If using early fusion, FiLM is the best conditioning method. For latent fusion and fusion everywhere (all), xAttn is a consistent choice, but FiLM may sometimes be better (and sometimes a lot worse). CAT in general should be avoided, which is consistent with the theoretical observation, that CAT is a special case of FiLM.

For SimVP, the best weather guiding method is latent fusion with FiLM. For PredRNN, the best method is early fusion with FiLM. This is likely due to the difference in treatment of the temporal axis. For SimVP, early fusion would merge all time steps, hence, latent fusion is a better choice. For PredRNN on the other hand, early fusion handles only a single timestep. 

\section{Contextformer Strengths and Limitations continued}
We show spatial extrapolation skills for more models in table~\ref{tab:extrapol2}.
\begin{table}
\centering
\begin{tabularx}{\columnwidth}{Xcccc}
\toprule
& \multicolumn{2}{c}{OOD-s} & \multicolumn{2}{c}{OOD-st} \\
Model & $R^2$ $\uparrow$ & RMSE $\downarrow$ & $R^2$ $\uparrow$ & RMSE $\downarrow$ \\
\midrule
Climatology & 0.50 & 0.15 & 0.56 & 0.19 \\
ConvLSTM & 0.47 & 0.17 & 0.52 & 0.16 \\
Earthformer & 0.47 &      0.15 &      0.47 &      0.16 \\
PredRNN & 0.54 & 0.15 & 0.58 & 0.15 \\
SimVP & 0.50 & 0.15 & 0.54 & 0.15 \\
Contextformer & 0.54 & 0.15 & 0.58 & 0.14 \\
\bottomrule
\end{tabularx}
\caption{Same as table~\ref{tab:extrapol}, but extended. Model skill at spatial (OOD-s) and spatio-temporal (OOD-st) extrapolation.}
\label{tab:extrapol2}
\end{table}

\begin{figure}
    \centering
    \begin{subfigure}[c]{0.49\columnwidth}
    \includegraphics[width=\textwidth]{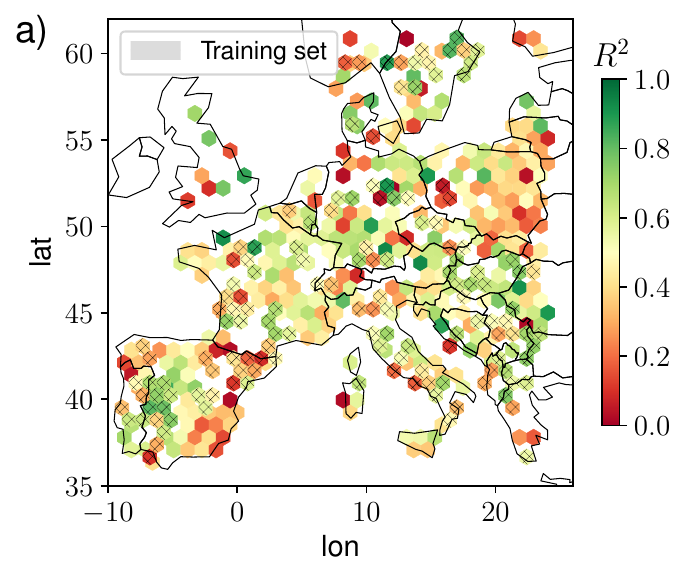}
    \end{subfigure}
    \begin{subfigure}[c]{0.49\columnwidth}
    \includegraphics[width=\textwidth]{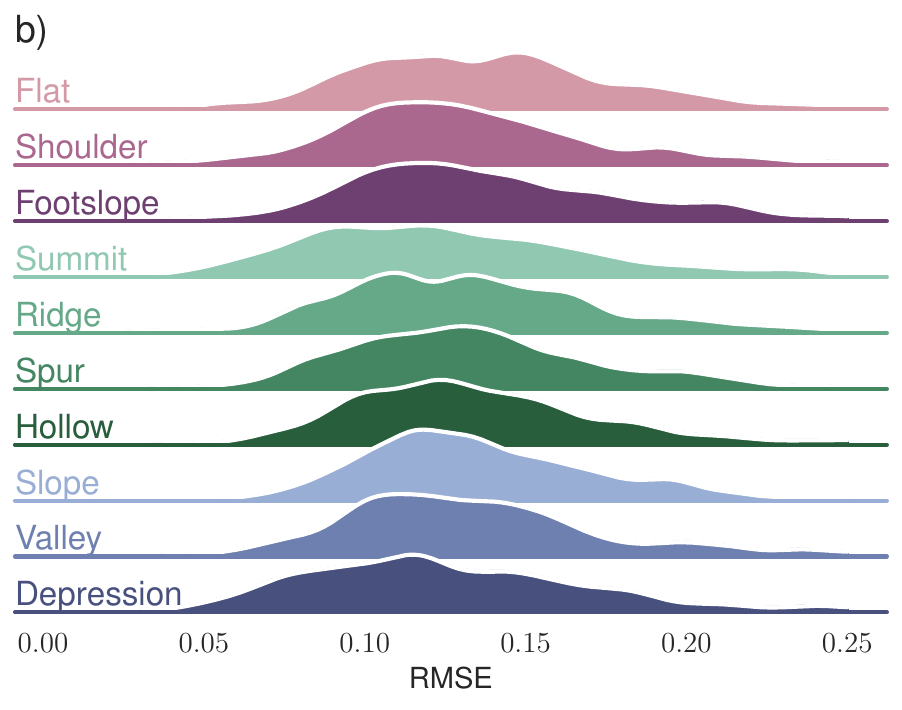}
    \end{subfigure}
    
    \caption{Panel a) shows a map of $R^2$ on OOD-t and OOD-st test sets and panel b) shows probability densities of RMSE per geomorphon. Both for Contextformer.}
    \label{fig:mapR2}
\end{figure}

Reassured by spatial extrapolation capabilities, we present a map of $R^2$ for the Contextformer in fig.~\ref{fig:mapR2}a. Cropland regions on the Iberian peninsula and in northern France, as well as forests in the Balkans are regions with great applicability of the model. For the former two, this may be explained by many training samples in those regions, for the last, it cannot. Grasslands and forests in Poland and highly heterogenous regions (mountains, near cities, near coasts) are more challenging for the model.

Geomorphons capture local terrain features, derived from first and second spatial derivatives of elevation. Fig.~\ref{fig:mapR2}b shows densities of RMSE of the ConvLSTM for different geomorphons from  the Geomorpho90m map \cite{amatulli.etal_2020}. Generally, the model performs well across all classes. Summits and Depressions, two rather extreme types, seem to be slightly easier to predict. Homogeneous terrain (red: flat, shoulder, footslope) has a larger tail towards high error. This may be as those regions are typically where there is a lot of anthropogenic activity, possibly leading to dynamics less covered by the predictors (harvest, clear-cut, etc.). 

\section{Performance per landcover type}
\begin{figure}
    \centering
    \includegraphics[width=\columnwidth]{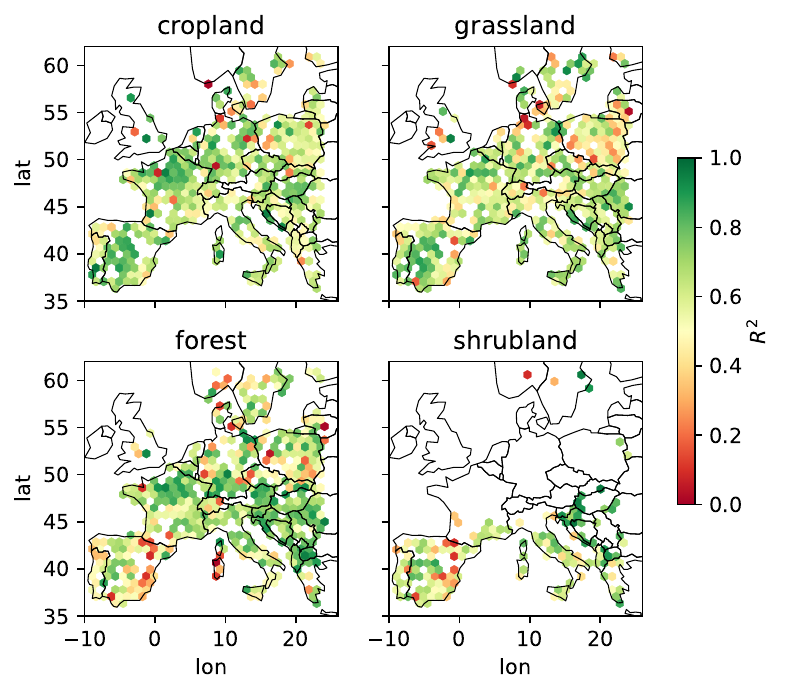}
    \caption{Model performance per landcover. Maps represent $R^2$ on OOD-t and OOD-st test sets of PredRNN.}
    \label{fig:map2}
\end{figure}

Fig.~\ref{fig:map2} shows the model performance per landcover type.

\section{Robustness of Outperformance Score}\label{sec:robustoutpf}
\begin{figure}[t]
    \centering
    \includegraphics[width=\columnwidth]{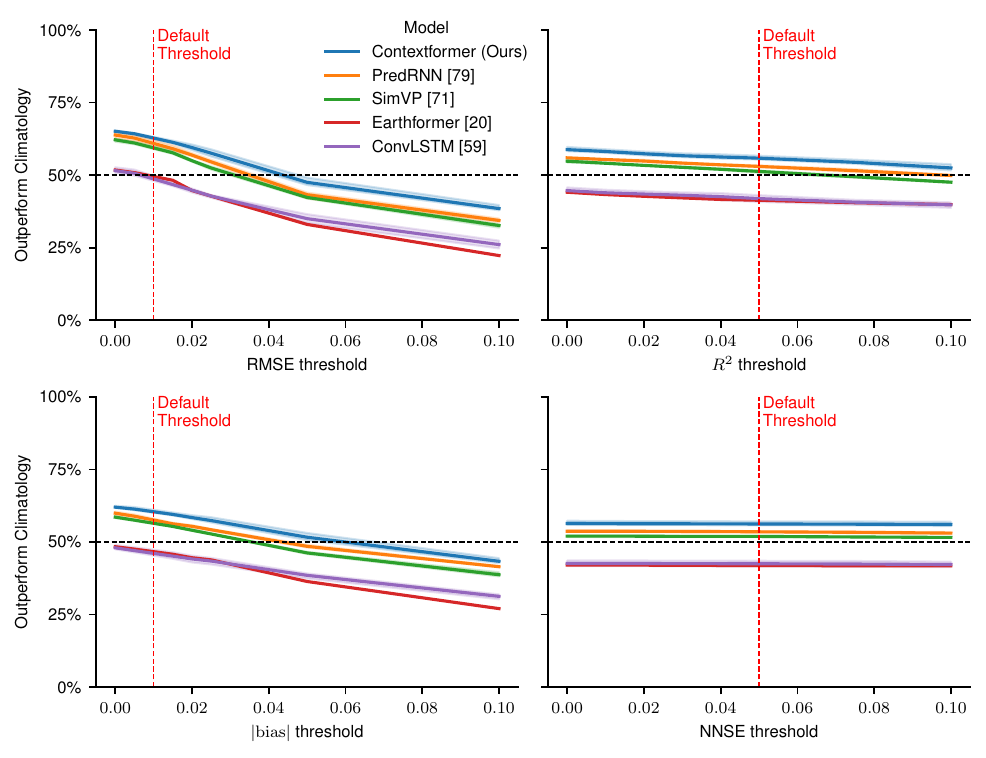}
    \caption{Outperformance score robustness (shown: marginals).}
    \label{fig:outpf}
\end{figure}

The choice of thresholds in the outperformance score (the percentage of samples where a model outperforms the climatology baseline by at least the threshold on at least 3 out of 4 metrics) is a heuristic. To assess its robustness, we re-evaluated five of our models over a wide range of possible threshold values. Fig.~\ref{fig:outpf} shows a consistency of the ranking, in particular our Contextformer outperforms all other models in all settings.

\section{Inference speed}\label{sec:speed}
Computing inference speed is highly platform and batch size dependent. To make it somewhat fair, we compare models by running 1024 samples on an A40 GPU (48GB), with the largest batch size (bs) fitting in memory, we perform 10 repetitions and report the mean and std. dev.: Contextformer $29.3$s{\footnotesize{$\pm 0.4$}} (bs 72), SimVP $6.7$s{\footnotesize{$\pm 0.8$}} (bs 96), PredRNN $16.2$s{\footnotesize{$\pm 0.2$}} (bs 512), ConvLSTM $37.1$s{\footnotesize{$\pm 1.8$}} (bs 256). For comparison predicting a single sample with one of the local time series models takes $>$1h on a single CPU.

\section{Downstream task: carbon monitoring}

Carbon monitoring is of great importance for climate change mitigation, especially in relation to nature-based solutions. The gross primary productivity (GPP) represents the amount of carbon that is taken up by plants through photosynthesis and subsequently stored. It is not directly observable. At a few hundred research stations around the world with eddy covariance measurement technology, it can be indirectly measured. For carbon monitoring, it would be beneficial to measure this quantity everywhere on the globe. It has been shown \cite{pabon-moreno.etal_2022} that Sentinel 2 NDVI is correlated to GPP measured with eddy covariance. We build on this correlation to show how our models could potentially be leveraged to give near real-time estimates of GPP and to study weather scenarios.

\begin{figure}
    \centering
    \includegraphics[width=0.95\columnwidth]{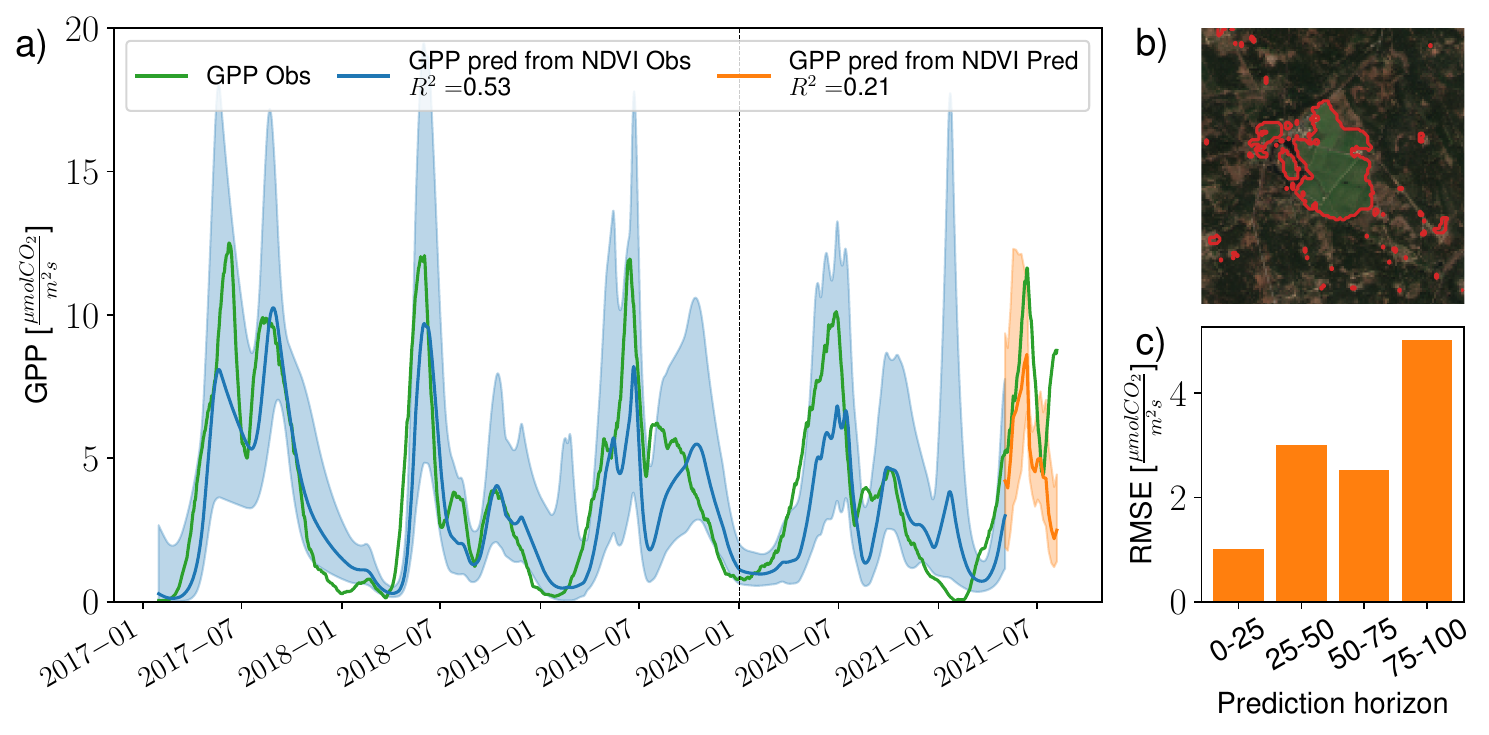}
    \caption{Panel a) shows timeseries of observed (green) and modeled GPP (blue from NDVI observations, orange from NDVI prediction). Panel b) shows a satellite image of the Grillenburg Fluxnet site with grassland boundaries in red. Panel c) shows the RMSE over prediction horizons.}
    \label{fig:gpp}
\end{figure}

Fig.~\ref{fig:gpp} compares modeled with observed GPP at the Fluxnet site Grillenburg (identifier DE-Gri) in eastern Germany distributed by ICOS \cite{degri_icos}. First, we fit a linear model between observed NDVI and GPP for the years 2017-2019. Here, interpolated grassland NDVI pixels (fig.~\ref{fig:gpp}b, inside red boundaries) are used. Next, we perform an out-of-sample analysis and find an $R^2 = 0.53$ for 2020-01 to 2021-04 (fig.~\ref{fig:gpp}a, blue line). Finally, we forecast GPP with our PredRNN model from May to July 2021(fig.~\ref{fig:gpp}a, orange line). The resulting forecast has decent quality at short prediction horizons, but low skill after 75 days (fig.~\ref{fig:gpp}c). These results show a way to leverage models from this paper for near real-time carbon monitoring. However, for application at scale, it is likely beneficial to use a more powerful GPP model (e.g. random forest \cite{pabon-moreno.etal_2022} or light-use efficiency \cite{bao.etal_2022}), fitted across many Fluxnet sites.

\end{document}